\documentclass[conference, 10pt]{IEEEtran}
\usepackage[T1]{fontenc}
\usepackage{graphicx}
\usepackage{booktabs}
\usepackage[misc]{ifsym}

\usepackage{layouts}

\usepackage{hyperref}
\usepackage{amsmath,amssymb,amsfonts}
\usepackage{algorithmic}
\usepackage{graphicx}
\usepackage{textcomp}
\usepackage{xcolor}
\usepackage{cleveref}
\usepackage{amsmath}
\usepackage{esvect}
\usepackage{xspace}
\usepackage{dsfont}
\usepackage{progressbar}

\usepackage{enumitem}

\usepackage[switch]{lineno}
\usepackage{url}

\usepackage{amsmath}

\usepackage{tikz}
\usetikzlibrary{positioning,decorations.markings,chains,arrows,arrows.meta,shadows,fadings,shapes,backgrounds,snakes,matrix,patterns,plotmarks,trees,mindmap, calc, tikzmark, fit}
\usepackage{tikzlings-squirrels}

\tikzset{
  treenode/.style={draw, circle, minimum size=1.5em, align=center}
}

\definecolor{class1}{rgb}{0.00392156862745098, 0.45098039215686275, 0.6980392156862745}
\definecolor{highlight}{rgb}{0.8352941176470589, 0.3686274509803922, 0.0}
\definecolor{outline}{rgb}{0.00784313725490196, 0.6196078431372549, 0.45098039215686275}
\definecolor{class0}{rgb}{0.8705882352941177, 0.5607843137254902, 0.0196078431372549}

\definecolor{optimal}{rgb}{0.00392156862745098, 0.45098039215686275, 0.6980392156862745}
\definecolor{forwardSquirrel}{rgb}{0.00784313725490196, 0.6196078431372549, 0.45098039215686275}
\definecolor{backwardSquirrel}{rgb}{0.8352941176470589, 0.3686274509803922, 0.0}

\definecolor{worst}{rgb}{0.8352941176470589, 0.3686274509803922, 0.0}
\definecolor{best}{rgb}{0.00784313725490196, 0.6196078431372549, 0.45098039215686275}

\tikzset{
    contour/.style={
      outline, 
      double,
      double distance=7mm,
      cap=round,
      thick,
    }
  }

  \tikzset{
  node split radius/.initial=1,
  scolor 1/.initial=class0,
  scolor 2/.initial=class0,
  scolor 3/.initial=class0,
  scolor 4/.initial=class0,
  scolor 5/.initial=class1,
  scolor 6/.initial=class1,
  scolor 7/.initial=class1,
  scolor 8/.initial=class1,
  node split half/.style={node split={#1,#1+180}},
  node split/.style={
    path picture={
      \tikzset{
        x=($(path picture bounding box.east)-(path picture bounding box.center)$),
        y=($(path picture bounding box.north)-(path picture bounding box.center)$),
        radius=\pgfkeysvalueof{/tikz/node split radius}}
        \fill[line join=round, draw, fill=\pgfkeysvalueof{/tikz/scolor 2}]
          (path picture bounding box.center)
          --++(0:\pgfkeysvalueof{/tikz/node split radius})
          arc[start angle=0, end angle=45] --cycle;
        \fill[line join=round, draw, fill=\pgfkeysvalueof{/tikz/scolor 1}]
          (path picture bounding box.center)
          --++(45:\pgfkeysvalueof{/tikz/node split radius})
          arc[start angle=45, end angle=90] --cycle;
        \fill[line join=round, draw, fill=\pgfkeysvalueof{/tikz/scolor 8}]
          (path picture bounding box.center)
          --++(90:\pgfkeysvalueof{/tikz/node split radius})
          arc[start angle=90, end angle=135] --cycle;
        \fill[line join=round, draw, fill=\pgfkeysvalueof{/tikz/scolor 7}]
          (path picture bounding box.center)
          --++(135:\pgfkeysvalueof{/tikz/node split radius})
          arc[start angle=135, end angle=180] --cycle;
        \fill[line join=round, draw, fill=\pgfkeysvalueof{/tikz/scolor 6}]
          (path picture bounding box.center)
          --++(180:\pgfkeysvalueof{/tikz/node split radius})
          arc[start angle=180, end angle=225] --cycle;
        \fill[line join=round, draw, fill=\pgfkeysvalueof{/tikz/scolor 5}]
          (path picture bounding box.center)
          --++(225:\pgfkeysvalueof{/tikz/node split radius})
          arc[start angle=225, end angle=270] --cycle;
        \fill[line join=round, draw, fill=\pgfkeysvalueof{/tikz/scolor 4}]
          (path picture bounding box.center)
          --++(270:\pgfkeysvalueof{/tikz/node split radius})
          arc[start angle=270, end angle=315] --cycle;
        \fill[line join=round, draw, fill=\pgfkeysvalueof{/tikz/scolor 3}]
          (path picture bounding box.center)
          --++(315:\pgfkeysvalueof{/tikz/node split radius})
          arc[start angle=315, end angle=360] --cycle;
} } }

\newcommand{\treeprogress}[4]{\begin{tikzpicture}
            \draw[draw, fill=white] (0,0) rectangle (0.15,0.3);
            \draw[draw, fill=white] (0.3,0) rectangle (0.45,0.3);
            \draw[draw, fill=white] (0.6,0) rectangle (0.75,0.3);
            
            \fill[black] (0,0.3-#1*0.15) rectangle (0.15,0.3); %
            \fill[black] (0.3,0.3-#2*0.15) rectangle (0.45,0.3); %
            \fill[black] (0.6,0.3-#3*0.15) rectangle (0.75,0.3); %

            \draw[draw] (0,0) rectangle (0.15,0.15);
            \draw[draw] (0.3,0) rectangle (0.45,0.15);
            \draw[draw] (0.6,0) rectangle (0.75,0.15);
        \end{tikzpicture}\\#4/8}

\definecolor{top}{RGB}{255,255,255}
\definecolor{bottom}{RGB}{0,0,0}

\newcommand{\esp}{ESP32\xspace}
\newcommand{\fullesp}{ESP32-S3-WROOM-2\xspace}
\newcommand{\epyc}{EPYC\xspace}
\newcommand{\fullepyc}{AMD EPYC 7742\xspace}

\newcommand{\optimal}{\emph{Optimal Order}\xspace}
\newcommand{\soptimal}{\emph{Optimal}\xspace}

\newcommand{\unoptimal}{\emph{Unoptimal Order}\xspace}

\newcommand{\hoptimal}{Optimal Order\xspace}
\newcommand{\squi}{\emph{Squirrel Order}\xspace}
\newcommand{\fsqui}{\emph{Forward Squirrel Order}\xspace}
\newcommand{\bsqui}{\emph{Backward Squirrel Order}\xspace}

\newcommand{\sfsqui}{\emph{Forward Squirrel}\xspace}

\newcommand{\sbsqui}{\emph{Backward Squirrel}\xspace}

\newcommand{\hsqui}{Squirrel Order\xspace}

\newcommand{\random}{\emph{Random Order}\xspace}

\newcommand{\depth}{\emph{Depth Order}\xspace}

\newcommand{\breadth}{\emph{Breadth Order}\xspace}

\newcommand{\qwyc}{\emph{QWYC Order}\xspace}

\newcommand{\qwycdepth}{\emph{QWYC Depth Order}\xspace}

\newcommand{\qwycbreadth}{\emph{QWYC Breadth Order}\xspace}

\newcommand{\qwycalgo}{\emph{QWYC Algorithm}\xspace}
\newcommand{\prune}{\emph{Prune Order}\xspace}
\newcommand{\prunebreadth}{\emph{Prune Breadth Order}\xspace}

\newcommand{\prunedepth}{\emph{Prune Depth Order}\xspace}

\newcommand{\normma}{normalized mean accuracy\xspace}
\newcommand{\nma}{NMA\xspace}

\begin{document}

\title{Jump Like A Squirrel: Optimized Execution Step Order for Anytime Random Forest Inference}

\author{
  \IEEEauthorblockN{1\textsuperscript{st} Daniel Biebert}
\IEEEauthorblockA{\textit{TU Dortmund University}\\
Dortmund, Germany \\
daniel.biebert@tu-dortmund.de}
\and
  \IEEEauthorblockN{2\textsuperscript{nd} Christian Hakert}
\IEEEauthorblockA{\textit{TU Dortmund University}\\
Dortmund, Germany \\
christian.hakert@tu-dortmund.de}
\and
  \IEEEauthorblockN{3\textsuperscript{rd} Kay Heider}
\IEEEauthorblockA{\textit{TU Dortmund University}\\
Dortmund, Germany \\
kay.heider@tu-dortmund.de}
\and
  \IEEEauthorblockN{4\textsuperscript{th} Daniel Kuhse}
\IEEEauthorblockA{\textit{TU Dortmund University}\\
Dortmund, Germany \\
daniel.kuhse@tu-dortmund.de}
\and
  \IEEEauthorblockN{5\textsuperscript{th} Sebastian Buschj{\"a}ger}
\IEEEauthorblockA{\textit{TU Dortmund University}\\
Dortmund, Germany \\
sebastian.buschjaeger@tu-dortmund.de}
\and
  \IEEEauthorblockN{6\textsuperscript{th} Jian-Jia chen}
\IEEEauthorblockA{\textit{TU Dortmund University}\\
Dortmund, Germany \\
jian-jia.chen@cs.tu-dortmund.de}

}

\maketitle              %

\begin{abstract}
  Due to their efficiency and small size, decision trees and random forests are popular machine learning models used for classification on resource-constrained systems. In such systems, the available execution time for inference in a random forest might not be sufficient for a complete model execution. Ideally, the already gained prediction confidence should be retained. An anytime algorithm is designed to be able to be aborted anytime, while giving a result with an increasing quality over time. Previous approaches have realized random forests as anytime algorithms on the granularity of trees, stopping after some but not all trees of a forest have been executed. However, due to the way decision trees subdivide the sample space in every step, an increase in prediction quality is achieved with every additional step in one tree. In this paper, we realize decision trees and random forest as anytime algorithms on the granularity of single steps in trees. This approach opens a design space to define the step order in a forest, which has the potential to optimize the mean accuracy. We propose the \optimal, which finds a step order with a maximal mean accuracy in exponential runtime and the polynomial runtime heuristics \fsqui and \bsqui, which greedily maximize the accuracy for each additional step taken down and up the trees, respectively.
  Our evaluation shows, that the \bsqui performs $\sim94\%$ as well as the \optimal and $\sim99\%$ as well as all other step orders.
\end{abstract}


\section{Introduction}

Decision trees and random forests are traditional, widely explored and widely applied machine learning models, which show strength in efficient solving especially of classification problems on tabular data \cite{DBLP:conf/nips/GrinsztajnOV22}.
They can be found in many specialized flavors, implementations and performance optimized versions \cite{DBLP:conf/icdm/BuschjagerCCM18,DBLP:journals/tecs/ChenSHBLLMC22}.
The model size can be reduced through methods such as pruning \cite{DBLP:journals/datamine/BuschjagerM23} to further reduce execution time.
However, while optimization and model size reduction can be used to meet an arbitrary execution time budget, these methods fail, if no suitable budget can be defined.
Without a guaranteed execution budget, it cannot be known how much to reduce the model size by.
Anytime algorithms aim to ensure that a result is returned in such situations.
The core idea of anytime algorithms is that regardless of when execution is stopped, a useful response is returned.
Furthermore, a longer runtime results in a higher quality response.
Solutions have been proposed, which enable a random forest to be executed as anytime algorithms \cite{DBLP:journals/jmlr/GrubbB12,DBLP:journals/jetc/WangGY21}.
Such implementations enable early stopping while retaining the full model.
They work by executing tree after tree in the forest and returning the prediction of all finished trees on execution stop.
These solutions therefore work on the granularity of trees.
All information gained in the currently executed tree is lost.

Due to the way decision trees are trained, the prediction confidence increases with each step.
More time spent evaluating data usually results in a more accurate classification.
By retaining information about a prediction in inner nodes, decision trees can be stopped on the granularity of single steps rather than full trees.
This results in no already gained information being lost whenever execution is stopped prematurely.

Without the limitations of having to execute full trees, it becomes possible to jump between trees before reaching a leaf.
This opens a design space to optimize the order in which to step through the trees towards higher accuracy.
We assume the probability of the execution to be stopped to be uniformly distributed over time.
We therefore want to optimize the mean accuracy across all steps.
In our first approach, called the \optimal, we find the step order that maximizes the mean accuracy on a sample set.
Building on that, we develop two heuristic approaches called \fsqui and \bsqui, which greedily maximize the accuracy for the next or previous step taken, respectively.

In short, we make the following contributions:
\begin{itemize}
    \item We enable general anytime execution of random forests with any combination of steps.
    \item We propose three mean accuracy optimizing anytime random forest step order generators, called \optimal, \fsqui, and \bsqui.
    \item We conduct experimental evaluation on a variety of data-sets with comparison to intuitive solutions.
\end{itemize}

\section{Related Work}
\label{sec:related}

Decision trees and random forests have been studied extensively in the literature. Much research has been conducted towards optimizing the learning of decision trees and random forests \cite{DBLP:conf/icdar/Ho95}. Ensemble pruning is a frequently used method to reduce the size of a random forest, while retaining or increasing its prediction accuracy (see \cite{DBLP:journals/datamine/BuschjagerM23} for a recent overview). The most common pruning methods are either ranking-based or greedy pruning methods. Ranking-based pruning methods assign a score to all decision trees in the forest, sort the trees by their scores, and select the top $n$ trees with the best scores \cite{DBLP:journals/ijon/GuoLLWGX18,DBLP:conf/kdd/LuWZB10,DBLP:conf/aaai/JiangLFW17,DBLP:conf/icml/MargineantuD97}. Greedy pruning strategies greedily maximize the accuracy by picking the best tree in each round from the remaining sub-forest until a specified number of trees have been selected \cite{DBLP:conf/icml/MargineantuD97,DBLP:conf/pkdd/LiYZ12}.

Specific efforts towards anytime decision tree and random forest learning have been made.
Esmeir et al. \cite{DBLP:journals/jmlr/EsmeirM07,DBLP:journals/ml/EsmeirM11} introduced a framework which trained decision trees, where training could be stopped at any point, while still receiving a trained tree. Further efforts have been made towards training trees which then are suitable to be used in the anytime context. Denninger et al. \cite{DBLP:conf/iros/DenningerT18} present a learning algorithm which incrementally trains random forests. Grubb et al. \cite{DBLP:journals/jmlr/GrubbB12} further created a framework to select predictors (such as decision trees) from a set of weak predictors to execute, which results in good performance when stopping after any of the selected predictors. A similar approach was developed by Wang et al. \cite{DBLP:journals/jetc/WangGY21}, which generates a tree sequence in a random forest as well as a threshold for an early stopping criterion.

These anytime strategies all either concentrate on an anytime learning algorithm, or executes the random forest on the granularity of trees rather than steps taken in trees. This means a tree has to be fully executed before it can be included in the prediction result. To the best of our knowledge, decision trees and random forests have not been adapted to predict any time on the granularity of a single step in one tree.

\section{Decision Trees and Random Forests}
\label{sec:background}

This section introduces decision trees and random forests \cite{DBLP:books/wa/BreimanFOS84,DBLP:journals/ml/Breiman01}, which are ensembles of decision trees, as well as adapts them to return a prediction at any time.

\subsection{Decision Trees}

Decision trees are trained and tested on a sample set. Each sample consists of a feature vector and is assigned a class. Given a data-set, it is usually split into a training set $\mathcal{S}_r$ and a test set $\mathcal{S}_t$. Similar to a validation set used for hyperparameter tuning, we further split the data-set into a third ordering set $\mathcal{S}_o$ to be used later \cite{Xu2018}.

Decision trees, in greater detail, are instances of directed acyclic graphs, where each node either has two follow-up nodes, denoted as the child nodes, or no follow-up nodes. Each tree must have exactly one root node. 
Nodes, which have child nodes, are denoted as inner nodes, while nodes without child nodes are denoted as leaf nodes. Inner nodes are associated with a pair of a feature index and a split value. By denoting the two child nodes of an inner node as the left child and right child, respectively, an execution rule can be defined.
During an inference of a sample, the next node is the left child, if the feature value at the feature index is less than or equal to the split value. Otherwise, the next node is the right child.
When execution reaches a leaf node, a prediction associated with the leaf node determines the classification result.

\subsection{Random Forests}
\label{subsec:background-forests}

A random forest is a collection of decision trees, where the predictions of the individual trees are combined into an overall prediction. Each leaf node returns a vector of probabilities, which associates each class with a probability. 
By summing up the probability vectors of all leaf nodes in all decision trees and choosing the class with the highest probability, the allover prediction is returned.
The method in this paper takes a random forest, described by the previous structure, as given. For the purposes of this paper, we used commonly used standard configurations of sklearn to train the forests \cite{DBLP:journals/corr/abs-1201-0490}.

\subsection{Predicting Any Time}
\label{subsec:pred-any-time}
For inference, fully trained decision trees usually only retains a prediction vector in leaf nodes. This, however, results in a strong limitation in prediction capabilities under anytime constraints, as it makes prediction only possible if a tree has reached a leaf node.
We therefore extend such designs by retaining the prediction vectors that typically arise during decision tree induction.
During training according to the CART algorithm \cite{DBLP:books/wa/BreimanFOS84}, at each node the training data-set is split along a split value in one dimension of the feature vector. This results in two sub-data-sets, one in the left child and one in the right child. Each node is therefore assigned its own sub-data-set of the full training set. Just as is done in leaf nodes, this is used to derive the prediction vector, which is the empirical probability vector of the classes in the sub-data-set. This information can be directly extracted from the CART algorithm \cite{DBLP:books/wa/BreimanFOS84}.
A combined prediction of multiple inner nodes can be achieved by applying the same method as explained in \Cref{subsec:background-forests}, which is to sum up all probability vectors of the nodes and returning the highest probable class. Any inner node can therefore also be used to return a prediction.

\section{Scheduling Within Random Forests}

Realizing random forests as an anytime algorithm on the granularity of steps opens a large design space.
The probability for an early abort is assumed to be equal at any step.
A step order should therefore maximize the mean accuracy.
In the following, the accuracy and all other metrics are calculated using the ordering set $\mathcal{S}_o$.

\subsection{Intuitive Step Orders}
\label{subsec:intuitive}

An intuitive way of executing the forest as an anytime algorithm is to follow the execution a standard realization usually follows. That is, execute one tree until a leaf is reached. Afterwards, the inference continues evaluating the next tree until all trees have reached a leaf.
As this step order goes the full depth in each tree, it is referenced as \depth in the following.
A similarly intuitive approach is to go through the trees layer by layer, rather than go the full depth in each tree. This type of step order is referenced as \breadth.

To define a sequence of the trees for the \depth and \breadth, we choose methods to sort the trees impact on accuracy from the literature.
One approach by Wang et al. \cite{DBLP:journals/jetc/WangGY21}, the \qwyc (Quit When You Can), assumes that some samples can be confidently classified after only some trees in the forest have been evaluated. The \qwycalgo generates a sequence in which to evaluate the trees.
The \qwycalgo is only designed for binary classification data-sets.

Another option is to take existing sequences that are used to prune a forest as explained in \Cref{sec:related} \cite{DBLP:conf/aaai/JiangLFW17,DBLP:conf/icml/MargineantuD97,DBLP:conf/pkdd/LiYZ12}.
In ranked pruning, each tree is given a score according to a metric and the highest scoring trees are chosen \cite{DBLP:conf/aaai/JiangLFW17}.
For our purposes, we sort all trees by their score and use this sequence.
When using greedy pruning, the trees are chosen iteratively, again subject to some metric (e.g., individual error) \cite{DBLP:conf/icml/MargineantuD97,DBLP:conf/pkdd/LiYZ12}. This can also be used to derive a step order by using the sequence the trees are chosen.
We call such a step order a \prune.
It should be noted that we retain all trees and only use the metrics to derive a sequence.
The resulting sequences can be used both with \depth or \breadth.

\subsection{\hoptimal}
\label{subsec:opt}

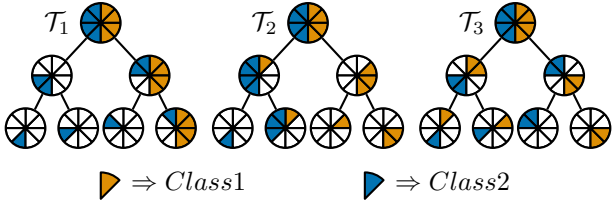
\begin{figure}
    \centering
    \begin{tikzpicture}[level distance=7mm, level 1/.style={sibling distance=13mm, thick}, level 2/.style={sibling distance=7mm, thick}, level 3/.style={sibling distance=4mm, thick}]

        \node [treenode, thick, node split] (r1){}
        child { node [treenode, node split,scolor 1=white,scolor 2=white,scolor 3=white,scolor 4=white,scolor 7=white,scolor 8=white,] (r1l) {}
                child { node [treenode, node split,scolor 1=white,scolor 2=white,scolor 3=white,scolor 4=white,scolor 6=white,scolor 7=white,scolor 8=white,] (r1ll) {}}
                child { node [treenode, node split,scolor 1=white,scolor 2=white,scolor 3=white,scolor 4=white,scolor 5=white,scolor 7=white,scolor 8=white,] (r1lr) {}}}
        child { node [treenode, node split,scolor 5=white,scolor 6=white,] (r1r) {}
                child { node [treenode, node split,scolor 1=white,scolor 2=white,scolor 3=white,scolor 4=white,scolor 5=white,scolor 6=white,scolor 8=white,] (r1rl) {}}
                child { node [treenode, node split,scolor 5=white,scolor 6=white,scolor 7=white,] (r1rr) {}}};

                \node [treenode, thick, right=of r1, xshift=12mm, node split] (r2){}
        child { node [treenode, node split,scolor 2=white,scolor 3=white,scolor 4=white,] (r2l) {}
                child { node [treenode, node split,scolor 1=white,scolor 2=white,scolor 3=white,scolor 4=white,scolor 6=white,scolor 7=white,scolor 8=white,] (r2ll) {}}
                child { node [treenode, node split,scolor 2=white,scolor 3=white,scolor 4=white,scolor 5=white,] (r2lr) {}}}
        child { node [treenode, node split,scolor 1=white,scolor 5=white,scolor 6=white,scolor 7=white,scolor 8=white,] (r2r) {}
                child { node [treenode, node split,scolor 1=white,scolor 3=white,scolor 4=white,scolor 5=white,scolor 6=white,scolor 7=white,scolor 8=white,] (r2rl) {}}
                child { node [treenode, node split,scolor 1=white,scolor 2=white,scolor 5=white,scolor 6=white,scolor 7=white,scolor 8=white,] (r2rr) {}}};

                \node [treenode, thick, right=of r2, xshift=12mm, node split] (r3){}
        child { node [treenode, node split,scolor 3=white,scolor 4=white,scolor 7=white,scolor 8=white,] (r3l) {}
                child { node [treenode, node split,scolor 2=white,scolor 3=white,scolor 4=white,scolor 6=white,scolor 7=white,scolor 8=white,] (r3ll) {}}
                child { node [treenode, node split,scolor 1=white,scolor 3=white,scolor 4=white,scolor 5=white,scolor 7=white,scolor 8=white,] (r3lr) {}}}
        child { node [treenode, node split,scolor 1=white,scolor 2=white,scolor 5=white,scolor 6=white,] (r3r) {}
                child { node [treenode, node split,scolor 1=white,scolor 2=white,scolor 3=white,scolor 4=white,scolor 5=white,scolor 6=white,] (r3rl) {}}
                child { node [treenode, node split,scolor 1=white,scolor 2=white,scolor 5=white,scolor 6=white,scolor 7=white,scolor 8=white,] (r3rr) {}}};

        \node[left=of r1, xshift=10mm] (t1) {$\mathcal{T}_1$};
        \node[left=of r2, xshift=10mm] (t1) {$\mathcal{T}_2$};
        \node[left=of r3, xshift=10mm] (t1) {$\mathcal{T}_3$};

        \node[treenode, thick, draw=white, outer sep=1mm, below=of r1, yshift=-7mm, xshift=0mm] (exampleclass0) {};
        \node[treenode, thick, draw=white, outer sep=1mm, below=of r3, yshift=-7mm, xshift=-20mm] (exampleclass1) {};

        \fill[line join=round, draw, thick, fill=class0] (exampleclass0.center)  --++(45:1em) arc[start angle=45, end angle=90, radius=1em] -- cycle;
        \fill[line join=round, draw, thick, fill=class1] (exampleclass1.center)  --++(45:1em) arc[start angle=45, end angle=90, radius=1em] -- cycle;

        \node[anchor=west] (exampleclass0text) at (exampleclass0.40) {$\Rightarrow Class 1$};
        \node[anchor=west] (exampleclass1text) at (exampleclass1.40) {$\Rightarrow Class 2$};

        \pgfdeclarelayer{bg}
        \pgfsetlayers{bg,main}

        \begin{pgfonlayer}{bg}

        \end{pgfonlayer}
    \end{tikzpicture}
    \caption{Example Forest with Marked Samples}
    \label{fig:forest-example}
\end{figure}

The intuitive step orders do not utilize the full potential of executing an anytime random forest on the granularity of single steps.
The forest can be traversed in any arbitrary step order.
$\frac{(d\cdot t)!}{d!^t}$ distinct step orders exist, where $d$ is the maximum depth and $t$ is the number of trees.
Each step order has a different mean accuracy.
When searching for an optimal step order, an exhaustive search is therefore not feasible.
However, the search space can be significantly reduced.
By interpreting the search space as a weighted directed acyclic graph, we can use Dijkstras algorithm \cite{DBLP:journals/nm/Dijkstra59} to find a step order with the maximum mean accuracy.
In the following, we explain how to construct the graph and use Dijkstras algorithm to find an optimal step order.
For this, we will use the example forest depicted in \Cref{fig:forest-example}. The example forest consists of 3 trees with depth 2. The ordering set $\mathcal{S}_o$ consists of 8 samples from two classes. The slices in each node show the samples reaching this node in an inference. The color indicates the class.

\begin{figure}
    \centering
    \begin{tikzpicture}[
            node/.style={rectangle, align=center, thick, draw, inner sep = 1mm, minimum size = 10mm},
        ]

        \node[node] (000) {\treeprogress{0}{0}{0}{4}};
        \node[node] (001) [below=of 000, yshift=3mm, xshift=-12mm] {\treeprogress{0}{0}{1}{4}};
        \node[node] (010) [below=of 000, yshift=3mm, xshift=0mm] {\treeprogress{0}{1}{0}{7}};
        \node[node] (100) [below=of 000, yshift=3mm, xshift=12mm] {\treeprogress{1}{0}{0}{6}};
        \node[node] (002) [below=of 010, yshift=3mm, xshift=-30mm] {\treeprogress{0}{0}{2}{6}};
        \node[node] (011) [below=of 010, yshift=3mm, xshift=-18mm] {\treeprogress{0}{1}{1}{7}};
        \node[node] (020) [below=of 010, yshift=3mm, xshift=-6mm] {\treeprogress{0}{2}{0}{7}};
        \node[node] (101) [below=of 010, yshift=3mm, xshift=6mm] {\treeprogress{1}{0}{1}{6}};
        \node[node] (110) [below=of 010, yshift=3mm, xshift=18mm] {\treeprogress{1}{1}{0}{7}};
        \node[node] (200) [below=of 010, yshift=3mm, xshift=30mm] {\treeprogress{2}{0}{0}{7}};
        \node[node] (012) [below=of 101, yshift=3mm, xshift=-42mm] {\treeprogress{0}{1}{2}{7}};
        \node[node] (021) [below=of 101, yshift=3mm, xshift=-30mm] {\treeprogress{0}{2}{1}{7}};
        \node[node] (102) [below=of 101, yshift=3mm, xshift=-18mm] {\treeprogress{1}{0}{2}{8}};
        \node[node] (111) [below=of 101, yshift=3mm, xshift=-6mm] {\treeprogress{1}{1}{1}{7}};
        \node[node] (120) [below=of 101, yshift=3mm, xshift=6mm] {\treeprogress{1}{2}{0}{7}};
        \node[node] (201) [below=of 101, yshift=3mm, xshift=18mm] {\treeprogress{2}{0}{1}{7}};
        \node[node] (210) [below=of 101, yshift=3mm, xshift=30mm] {\treeprogress{2}{1}{0}{7}};
        \node[node] (022) [below=of 111, yshift=3mm, xshift=-30mm] {\treeprogress{0}{2}{2}{7}};
        \node[node] (112) [below=of 111, yshift=3mm, xshift=-18mm] {\treeprogress{1}{1}{2}{7}};
        \node[node] (121) [below=of 111, yshift=3mm, xshift=-6mm] {\treeprogress{1}{2}{1}{7}};
        \node[node] (202) [below=of 111, yshift=3mm, xshift=6mm] {\treeprogress{2}{0}{2}{8}};
        \node[node] (211) [below=of 111, yshift=3mm, xshift=18mm] {\treeprogress{2}{1}{1}{7}};
        \node[node] (220) [below=of 111, yshift=3mm, xshift=30mm] {\treeprogress{2}{2}{0}{7}};
        \node[node] (122) [below=of 202, yshift=3mm, xshift=-18mm] {\treeprogress{1}{2}{2}{7}};
        \node[node] (212) [below=of 202, yshift=3mm, xshift=-6mm] {\treeprogress{2}{1}{2}{8}};
        \node[node] (221) [below=of 202, yshift=3mm, xshift=6mm] {\treeprogress{2}{2}{1}{7}};
        \node[node] (222) [below=of 212, yshift=3mm, xshift=0mm] {\treeprogress{2}{2}{2}{8}};
        \draw[->] (000.300) to (100.90);
        \draw[->, very thick, optimal] (000.262.5) to (010.97.5);
        \draw[->, very thick, forwardSquirrel] (000.277.5) to (010.82.5);
        \draw[->, very thick, backwardSquirrel] (000.240) to (001.90);
        \draw[->] (001.300) to (101.105);
        \draw[->] (001.270) to (011.105);
        \draw[->, very thick, backwardSquirrel] (001.240) to (002.90);
        \draw[->, very thick, backwardSquirrel] (002.285) to (102.105);
        \draw[->] (002.255) to (012.105);
        \draw[->, very thick, optimal] (010.300) to (110.105);
        \draw[->] (010.270) to (020.90);
        \draw[->, very thick, forwardSquirrel] (010.240) to (011.75);
        \draw[->] (011.300) to (111.120);
        \draw[->] (011.270) to (021.105);
        \draw[->, very thick, forwardSquirrel] (011.240) to (012.75);
        \draw[->] (012.285) to (112.120);
        \draw[->, very thick, forwardSquirrel] (012.255) to (022.105);
        \draw[->] (020.285) to (120.105);
        \draw[->] (020.255) to (021.75);
        \draw[->] (021.285) to (121.120);
        \draw[->] (021.255) to (022.75);
        \draw[->, very thick, forwardSquirrel] (022.270) to (122.120);
        \draw[->] (100.300) to (200.90);
        \draw[->] (100.270) to (110.75);
        \draw[->] (100.240) to (101.75);
        \draw[->] (101.300) to (201.105);
        \draw[->] (101.270) to (111.90);
        \draw[->] (101.240) to (102.75);
        \draw[->, very thick, backwardSquirrel] (102.285) to (202.105);
        \draw[->] (102.255) to (112.90);
        \draw[->, very thick, optimal] (110.300) to (210.105);
        \draw[->] (110.270) to (120.75);
        \draw[->] (110.240) to (111.60);
        \draw[->] (111.300) to (211.120);
        \draw[->] (111.270) to (121.90);
        \draw[->] (111.240) to (112.60);
        \draw[->] (112.285) to (212.120);
        \draw[->] (112.255) to (122.90);
        \draw[->] (120.285) to (220.105);
        \draw[->] (120.255) to (121.60);
        \draw[->] (121.285) to (221.120);
        \draw[->] (121.255) to (122.60);
        \draw[->, very thick, forwardSquirrel] (122.270) to (222.120);
        \draw[->] (200.285) to (210.75);
        \draw[->] (200.255) to (201.75);
        \draw[->] (201.285) to (211.90);
        \draw[->] (201.255) to (202.75);
        \draw[->, very thick, backwardSquirrel] (202.270) to (212.90);
        \draw[->] (210.285) to (220.75);
        \draw[->, very thick, optimal] (210.255) to (211.60);
        \draw[->] (211.285) to (221.90);
        \draw[->, very thick, optimal] (211.255) to (212.60);
        \draw[->, very thick, optimal] (212.262.5) to (222.97.5);
        \draw[->, very thick, backwardSquirrel] (212.277.5) to (222.82.5);
        \draw[->] (220.270) to (221.60);
        \draw[->] (221.270) to (222.60);
        \pgfdeclarelayer{bg}
        \pgfsetlayers{bg,main}
        \begin{pgfonlayer}{bg}
            \filldraw[optimal!40][] (000.north west) -- (000.70) -- (000.200) -- cycle;
            \filldraw[forwardSquirrel!40][] (000.70) -- (000.north east) -- (000.20) -- (000.250) -- (000.south west) -- (000.200) -- cycle;
            \filldraw[backwardSquirrel!40][] (000.south east) -- (000.250) -- (000.20) -- cycle;
            \filldraw[optimal!40][] (010.north west) -- (010.north east) -- (010.south west) -- cycle;
            \filldraw[forwardSquirrel!40][] (010.north east) -- (010.south east) -- (010.south west) -- cycle;
            \filldraw[backwardSquirrel!40][] (001.north west) -- (001.north east) -- (001.south east) -- (001.south west) -- cycle;
            \filldraw[forwardSquirrel!40][] (011.north west) -- (011.north east) -- (011.south east) -- (011.south west) -- cycle;
            \filldraw[backwardSquirrel!40][] (002.north west) -- (002.north east) -- (002.south east) -- (002.south west) -- cycle;
            \filldraw[optimal!40][] (110.north west) -- (110.north east) -- (110.south east) -- (110.south west) -- cycle;
            \filldraw[backwardSquirrel!40][] (102.north west) -- (102.north east) -- (102.south east) -- (102.south west) -- cycle;
            \filldraw[forwardSquirrel!40][] (012.north west) -- (012.north east) -- (012.south east) -- (012.south west) -- cycle;
            \filldraw[optimal!40][] (210.north west) -- (210.north east) -- (210.south east) -- (210.south west) -- cycle;
            \filldraw[forwardSquirrel!40][] (022.north west) -- (022.north east) -- (022.south east) -- (022.south west) -- cycle;
            \filldraw[backwardSquirrel!40][] (202.north west) -- (202.north east) -- (202.south east) -- (202.south west) -- cycle;
            \filldraw[optimal!40][] (211.north west) -- (211.north east) -- (211.south east) -- (211.south west) -- cycle;
            \filldraw[forwardSquirrel!40][] (122.north west) -- (122.north east) -- (122.south east) -- (122.south west) -- cycle;
            \filldraw[optimal!40][] (212.north west) -- (212.north east) -- (212.south west) -- cycle;
            \filldraw[backwardSquirrel!40][] (212.north east) -- (212.south east) -- (212.south west) -- cycle;
            \filldraw[optimal!40][] (222.north west) -- (222.70) -- (222.200) -- cycle;
            \filldraw[forwardSquirrel!40][] (222.70) -- (222.north east) -- (222.20) -- (222.250) -- (222.south west) -- (222.200) -- cycle;
            \filldraw[backwardSquirrel!40][] (222.south east) -- (222.250) -- (222.20) -- cycle;
        \end{pgfonlayer}

        \node[align=left, anchor=north west, inner sep = 0mm] (orders) at ([yshift=0pt]000.north -| 012.west) {\textcolor{optimal}{\soptimal}\\\textcolor{forwardSquirrel}{\sfsqui}\\\textcolor{backwardSquirrel}{\sbsqui}};

        \node[align=center, anchor=north east, inner sep = 0mm] (steps) at ([yshift=0pt]000.north -| 210.east) {\begin{tikzpicture}
                \fill[black] (0,0.3) rectangle (0.15,0.3); %
                \draw[draw] (0,0) rectangle (0.15,0.3);
                \draw[draw] (0,0) rectangle (0.15,0.15);
                \node[anchor=west, inner sep = 0mm] at (0.3,0.15) {0 Steps};

                \fill[black] (0,0.15-.5) rectangle (0.15,0.3-.5); %
                \draw[draw] (0,-.5) rectangle (0.15,0.3-.5);
                \draw[draw] (0,0-.5) rectangle (0.15,0.15-.5);
                \node[anchor=west, inner sep = 0mm] at (0.3,0.15-.5) {1 Step};

                \fill[black] (0,0-1) rectangle (0.15,0.3-1); %
                \draw[draw] (0,-1) rectangle (0.15,0.3-1);
                \draw[draw] (0,0-1) rectangle (0.15,0.15-1);
                \node[anchor=west, inner sep = 0mm] at (0.3,0.15-1) {2 Steps};
            \end{tikzpicture}
        };

        \node[align=center, anchor=south west, inner sep = 0mm] (trees) at ([yshift=0pt]222.south -| 012.west) {\begin{tikzpicture}
                \fill[black] (-.375 + 0,0.3) rectangle (-.375 + 0.15,0.3); %
                \fill[black] (-.375 + 0.3,0.3-0.15) rectangle (-.375 + 0.45,0.3); %
                \fill[black] (-.375 + 0.6,0.3-0.3) rectangle (-.375 + 0.75,0.3); %

                \draw[draw] (-.375 + 0,0) rectangle (-.375 + 0.15,0.3);
                \draw[draw] (-.375 + 0.3,0) rectangle (-.375 + 0.45,0.3);
                \draw[draw] (-.375 + 0.6,0) rectangle (-.375 + 0.75,0.3);

                \draw[draw] (-.375 + 0,0) rectangle (-.375 + 0.15,0.15);
                \draw[draw] (-.375 + 0.3,0) rectangle (-.375 + 0.45,0.15);
                \draw[draw] (-.375 + 0.6,0) rectangle (-.375 + 0.75,0.15);

                \draw[->] (-0.3, 0) to (-.75,-.75);
                \draw[->] (0, 0) to (0,-.75);
                \draw[->] (0.3, 0) to (.75,-.75);

                \node[anchor=north, inner sep = 0mm] at (-.75,-.75) {$\mathcal{T}_1$};
                \node[anchor=north, inner sep = 0mm] at (0,-.75) {$\mathcal{T}_2$};
                \node[anchor=north, inner sep = 0mm] at (.75,-.75) {$\mathcal{T}_3$};

            \end{tikzpicture}
        };

        \node[align=center, anchor=south east, inner sep = 0mm] (trees) at ([yshift=0pt]222.south -| 210.east) {\begin{tikzpicture}
                \node[anchor=center]  {6/8};

                \draw[->] (-.2, -.1) to (-.75,-.75);
                \draw[->] (.2, -.1) to (.75,-.75);

                \node[anchor=north, inner sep = 0mm, align=center] at (-.75,-.75) {Correct\\Samples};
                \node[anchor=north, inner sep = 0mm, align=center] at (.75,-.75) {Total\\Samples};

            \end{tikzpicture}
        };

    \end{tikzpicture}
    \caption{Graph of Example Forest}
    \label{fig:graph-example}
\end{figure}
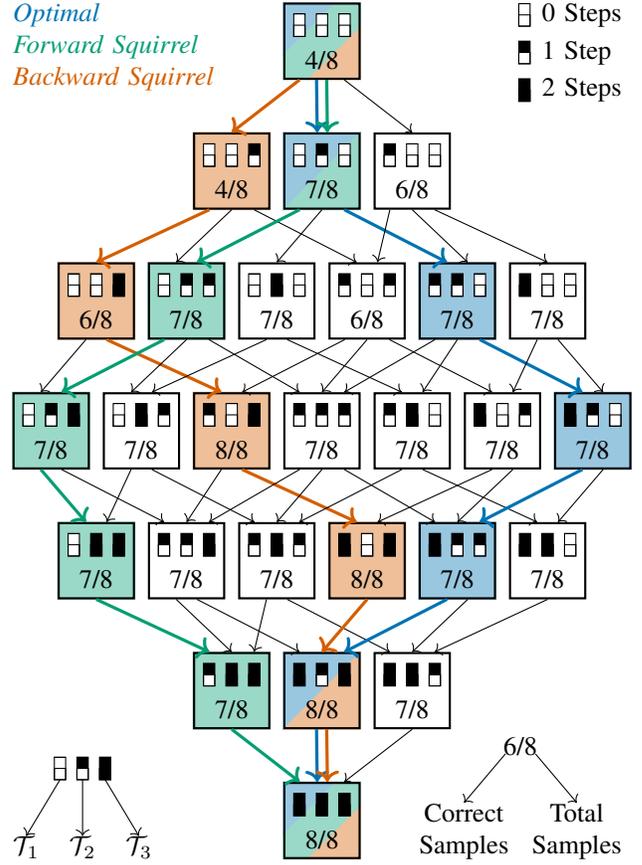

We first build the graph.
The graph for the example forest is shown in \Cref{fig:graph-example}.
We model the vertices of the graph as the possible states the forest can be in during inference.
A state is defined by the number of steps taken in each of the trees in the forest.
In \Cref{fig:graph-example}, the left most vertex shows the state with the first tree having taken zero steps, the second tree one step and the third tree two steps.
All vertices are connected with outgoing edges to vertices that model one additional step being taken in one tree.
In the example, the left most vertex connects to either the first or second tree having taken one additional step.
The weight of an edge is determined by the accuracy of state it leads into.
The accuracy of a state is calculated using the ordering set $\mathcal{S}_o$.
In the example, the state of the left most vertex correctly classifies seven of the eight samples.
All edges leading into this vertex therefore have a weight of $\frac{7}{8}$.
The graph models all distinct ways to step through the tree.

Maximizing the mean accuracy is the same as minimizing the mean inaccuracy.
As all step orders have the same number of steps we can simply use the sum of inaccuracies instead of the mean inaccuracy. Our goal can be rewritten as minimizing the sum of inaccuracies of all steps.

To find the optimal order, we now use Dijkstras algorithm \cite{DBLP:journals/nm/Dijkstra59} with the starting vertex being no steps taken and the ending vertex being all steps taken.
The weights are changed to the inverse of the accuracy, to minimize the inaccuracy.
As Dijkstras algorithm finds the minimal path through a graph, we are guaranteed to have found the step order with the minimum mean inaccuracy and therefore the highest mean accuracy.
The \optimal for the example is marked blue in \Cref{fig:graph-example}.
The mean accuracy is $\frac{6}{7}$ or $\sim 86\%$.
The size of the graph is $(d+1)^t$. The Dijkstra algorithm therefore has a worst-case runtime of $\mathcal{O}((d+1)^t \cdot \log((d+1)^t))$ \cite{DBLP:journals/nm/Dijkstra59}.

\subsection{\hsqui}
\label{subsec:squi}

While the \optimal results in a guaranteed maximal mean accuracy in the ordering set, it cannot be feasibly calculated for forests of increasing sizes. We therefore introduce the heuristic \squi, with a polynomial runtime. We distinguish the \fsqui and \bsqui, which both operate on the same principle, with the difference being the direction the step order is generated in.

The core idea of the \squi is to do a greedy depth first search through the graph.
The \fsqui starts at the initial state and finds the next state with the highest accuracy.
The \fsqui continues from this state and again, finds the following state with the highest accuracy.
This is repeated iteratively until the final state is reached.
In \Cref{fig:graph-example}, the \fsqui is marked in green.
The first step is chosen to be in the second tree, as this results in the highest accuracy of $\frac{7}{8}$.
The \bsqui operates very similarly, only starting from the final state and going backwards in the tree until the initial state is reached.
The \bsqui is marked in red.
When going backwards in the example it chooses the second tree to be the last one to take a step, as this results in the highest accuracy in the second to last step of $\frac{8}{8}$.

While both variants of the \squi operate on the graph on a logical level, crucially, the full graph does not need to be constructed.
Rather, the depth first search finds the $t$ next states.
The accuracy only has to be calculated for these states.
Furthermore, the graph does not have to be stored, reducing the memory footprint.
The runtime is therefore $\mathcal{O}(d\cdot t^2)$\footnote{Assuming calculating the accuracy is constant. In practice the accuracy calculation scales with the number of samples and trees.}.
It should be noted, that both the generation of the \optimal and the \squi variants only need to be done once, before inference. Once the orders are generated, they remain unchanged and do not increase inference runtime.

\section{Implementation of Anytime Random Forests}
While the determination of the step order through the forest under anytime constraints is conducted on a logical perspective of a random forest, the actual technical ability to stop the execution of a decision tree prior to the normal termination is discussed in this section. 
Decision trees are commonly implemented as native trees \cite{DBLP:journals/tkde/AsadiLV14}. In native trees, the tree data is stored in an array and a loop iterates over the data, forming the inference algorithm.
Two conditions are necessary for an anytime implementation of a random forest:
\begin{enumerate}
    \item Ability to leave and return to a tree before completion \label{cond:leave}
    \item Current state needs to be translatable to a prediction \label{cond:pred}
\end{enumerate}

To realize these conditions in native trees, the state of the inference is kept by means of an index array, which stores the index of the current node in each tree. The inference is realized by iterating over an array encoding the step order, taking a step in the next tree and updating the node index of that tree.
To get the prediction in case of an early abort, the index array can be used to load the appropriate prediction vectors of all trees and combining them to the allover prediction. The resulting implementation therefore realizes a tight loop over the step order, advancing one tree for every step. It further enables the ability to get a prediction at any time.

A competing random forest implementation paradigm is if-else trees \cite{DBLP:journals/tkde/AsadiLV14}. In if-else trees, the tree structure is encoded in the source code itself, realizing split decisions as if-else statements. If-else trees can also be realized to be an anytime implementation. However, much larger effort compared to native trees is required, as leaving and returning to a tree means jumping back to the appropriate location in the code. An optimized anytime random forest realization implementing if-else trees is therefore an interesting topic for future work.

\section{Evaluation}
\label{sec:eval}

To evaluate both the validity of executing a random forest as an anytime algorithm on the granularity of single steps as well as the accuracy performance of the different step orders, experiments were run. 9 data-sets were chosen from the UCI Machine Learning Repository \cite{Lichman:2013} and trained using sklearn \cite{DBLP:journals/corr/abs-1201-0490}. The data-sets are adult, covertype, letter, magic, mnist, satlog, sensorless-drive, spambase and wearable-body-postures.
The data-sets are randomly sampled into three sub-sets, train (50\%), ordering (25\%) and test (25\%). 
The train sub-set is used to train the random forest.
The ordering sub-set is used to generate the step orders.
The test sub-set is used for inference and accuracy calculation in the evaluation.
The experiments were repeated with 5 different random seeds for every combination of number of trees and maximum depth for every data-set.
The experiments were run on two systems, one equipped with an \fullepyc (\epyc for brevity) with 251 GiB of memory and the other with an \fullesp (\esp for brevity).

Experiments were run with the following step orders:

\begin{itemize}
    \item \optimal.
    \item \fsqui and \bsqui{}.
    \item \prunedepth and \prunebreadth. Specifically, we used \emph{individual error} (IE) and \emph{error ambiguity} (EA) \cite{DBLP:conf/aaai/JiangLFW17} for rank-based pruning and reduced error (RE) and drep (D) \cite{DBLP:conf/icml/MargineantuD97,DBLP:conf/pkdd/LiYZ12} for greedy pruning.
    \item \qwycdepth and \qwycbreadth  \cite{DBLP:journals/jetc/WangGY21}.
\end{itemize}

We further use two na\"ive step orders for comparison:
\begin{itemize}
    \item \unoptimal, which works the same as \optimal but minimizes the mean accuracy.
    \item \random, which randomly chooses the next step.
\end{itemize}

It should be noted that all step orders are implemented to use early abort on the granularity of single steps.

\subsection{Relation Between Time and Steps}
\label{subsec:eval-time}
In the previous parts of this paper, steps within single trees of a forest are considered as the central unit for progress.
The unit of steps, although allowing a precise relation to the structure of a random forest, does not have a direct relation to the execution time. In the context of an anytime algorithm, abortion of the algorithm could be triggered after a certain amount of time. The relation between steps in a random forest and execution time is discussed in this subsection.

\begin{figure}
    \centering
    \includegraphics[width=.85\linewidth, trim={0 15pt 0 15pt}]{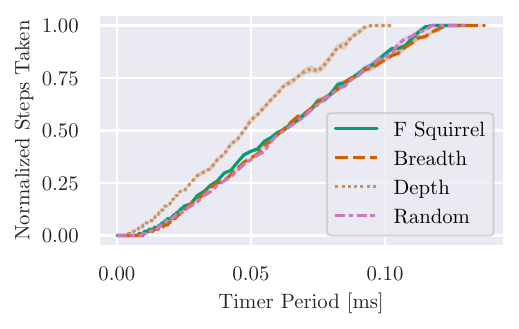}
    \caption{Expiry Time vs. Steps - adult (10 Trees, Depth 10)}
    \label{fig:timesteps}
\end{figure}

In order to estimate the relation of  steps  with execution time, we performed an experiment on the \esp, representing an embedded platform, where we configured a timer to interrupt the random forest execution after a configured expiry period. The timer was sourced with a high precision clock. This timer interrupt was considered to be an external event, causing the anytime abort of the random forest. We recorded how many steps the forest executed until the timer interrupt and report this as a normalized number to the total amount of steps for different configured timer periods. \Cref{fig:timesteps} depicts the configured timer periods on the x-axis with the corresponding normalized number of executed steps on the y-axis for a selection of step orders.
All experiments were repeated 10 times. The depicted configuration is trained on the adult data-set with 10 trees and a maximal depth of 10.

The experiments show a largely linear relation between steps and time in the average case, allowing the abstraction of steps as a measure of time. While small differences in execution speed of the different step orders can be observed, these differences do not affect the \optimal or \squi generation, as two steps are never faster than one step.

Conducting evaluation with a major focus on execution time leads to noise-prone results. Although timing is constant on average, it is not considering a single execution. Comparing two experiments with the same timer period might report two widely different progress states. This makes evaluation of the forest trend over increasing progress impossible. The rest of this evaluation studies the forest over the number of executed steps as a measure of progress.

\subsection{Step Order Generation Runtime}
\label{subsec:run-time}

\begin{figure}
    \centering
    \includegraphics[width=\linewidth, trim={0 15pt 0 15pt}]{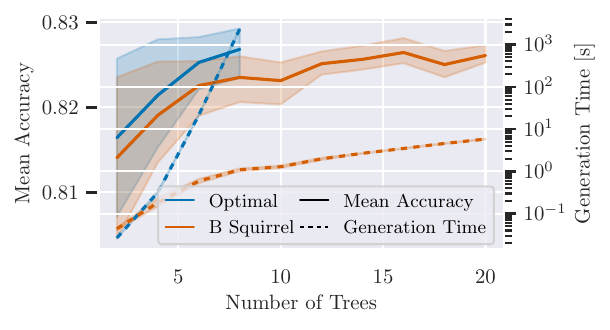}
    \caption{Step Order Runtime and Accuracy - adult (Depth 8)}
    \label{fig:runtimes}
\end{figure}

To analyze the step order generation runtime, we conducted an experiment on the adult data-set where we measured the time to generate the step orders. We kept the maximum depth at 8 and varied the number of trees from 2 to 20 in steps of 2. We generated the \optimal and the \bsqui on the \epyc, parallelizing the generation as much as possible. In \Cref{fig:runtimes} we plot the number of trees on the x-axis.
We plot the mean accuracy on the left y-axis and the generation runtime on the right y-axis in a logarithmic scale.

The \optimal exhibits exponential generation runtime. After 8 trees, the runtime and memory footprint became too large to generate the \optimal. The \bsqui shows significantly lower runtime with comparable mean accuracy (we will look at accuracy more closely later).
Due to its significantly lower runtime, the \bsqui can be used to generate a step order for much larger and deeper forests.
It is worth noting again, that the step order only needs to be generated once, prior to inference. It does not impact the runtime of the random forest inference.

\subsection{Relation Between Steps and Accuracy}
\label{subsec:eval-accuray}

For an in depth evaluation of the step orders impact on accuracy, we ran experiments where the number of steps until the inference was aborted were varied. Experiments with all step orders were run where we varied the number of trees and maximum depth across 4-7 with all combinations run. To evaluate all step orders except the \optimal on larger forests, additional experiments without the \optimal were run with number of trees 5, 10, and 20 and maximum depth of 2, 5, 10, and 20.

\begin{figure}
    \centering
        \includegraphics[width=.85\linewidth, trim={0 15pt 0 15pt}]{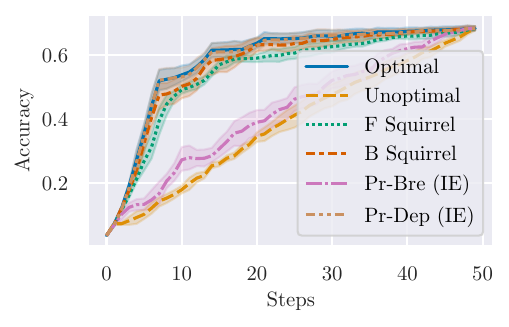}
    \caption{Steps vs. Accuracy - letter (7 Trees, Depth 7)}
    \label{fig:acc}
\end{figure}

First, we compare the relationship of number of steps to accuracy for a specific configuration. \Cref{fig:acc} shows the results for forests with 7 trees and a maximal depth of 7, trained on the letter data-set. The x-axis shows the number of steps and the y-axis the resulting accuracy. For now, we only display a selection of step orders.

Naturally, all step orders start from and converge to the same accuracy. For most step orders the accuracy increases quickly with only a few steps taken. 
Both \squi versions, the \optimal, and \prunedepth increase accuracy most quickly. The \unoptimal and the \prunedepth show a shallower slope.

The results show, that random forests are well suited to be executed as an anytime algorithm. The accuracy increases with steps taken. That means that taking another step usually improves the accuracy, and spending more time executing a forest results in a better prediction.
The chosen step order has a large impact on the slope of the accuracy curve.

\begin{figure}
    \centering
    \includegraphics[width=\linewidth, trim={0 15pt 0 15pt}]{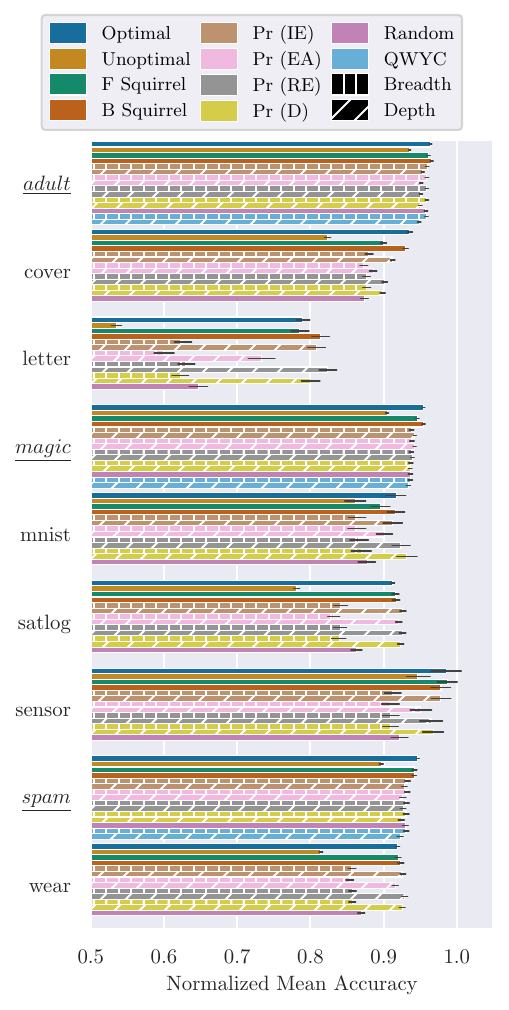}
    \caption{Average Normalized Mean Accuracy - All Data-Sets}
    \label{fig:norm}
\end{figure}

To allow comparison of all step orders, the following uses the mean accuracy as the metric to compare. We add a normalization step to make different configurations comparable. We divide the mean accuracy by the product of the total number of steps and the final accuracy, i.e. achieving the final accuracy at every step. The result is the \normma (\nma).
A higher \nma is better than a lower.
We plot the \nma for all run configurations in \Cref{fig:norm}. The x-axis shows the \nma. The y-axis shows the different data-sets. All \depth and \breadth variants of the same tree sequence are colored in the same color, marked with diagonal and vertical stripes, respectively.
The binary classification data-sets are underlined.

The \nma differs largely between data-sets.
The nature of the slope is therefore not only step order dependent, but also data-set and forest dependent.
The \optimal performs best and the \unoptimal performs worst in most data-sets.
In some cases, other step orders perform better than the \optimal.
While the \optimal results in an optimal mean accuracy for the ordering set $\mathcal{S}_o$, it may perform different on the test data-set $\mathcal{S}_t$.
The \bsqui performs similarly well compared to the \optimal and usually outperforms the \fsqui.

The depth variants perform significantly better than the breadth variants in non-binary classification data-sets.
The opposite can be observed in binary classification data-sets.
Furthermore, the overall range in the \nma is smaller in binary classification data-sets.
The difference between the two types of data-sets likely stems from the need for more steps for a confident classification in non-binary classification data-sets.
These types of data-sets therefore prefer depth variants, as only at a certain depth all classes are distinguished.

The \optimal and the \bsqui perform well regardless of the data-set.
They can confidently find a step order that achieves a high \nma.
The \bsqui performs similarly well compared to the \optimal.
It does so while generating the step order significantly faster.
In experiments with the \optimal, on average the \optimal achieved a \nma $\sim97\%$ as high as the highest achieved \nma. The \bsqui achieved $\sim94\%$ as high a \nma.
In experiments without the \optimal, the \bsqui achieved a \nma $\sim99\%$ as high as the highest \nma.

\section{Conclusion}

In this paper, we showed, that by retaining the prediction information in the inner nodes, a decision tree can be executed as an anytime algorithm. We further showed, that it is possible to jump through random forests on the granularity of single steps in a single tree, while retaining the ability to return a meaningful prediction. This opened an optimization problem of finding a step order through the trees for which the mean accuracy is maximized. We showed multiple intuitive strategies to derive step orders from solutions to similar problems in the literature. We developed, how to derive a step order that maximizes mean accuracy on the ordering data-set (\optimal). 
We reduce the order generation runtime to polynomial with the heuristic \squi, which also tries to maximize mean accuracy. It can be applied both in the direction of the leaves (\fsqui) and towards the roots (\bsqui). We showed that an anytime implementation can be realized efficiently.
The evaluation showed, that the polynomial runtime of the \squi enables the step order generation of much larger forests in shorter amount of time.
We further saw, that the prediction accuracy of an anytime random forests increases with additional steps taken, showing the validity of executing a random forest on the granularity of single steps.
The evaluation lastly showed, that while different intuitive strategies performed well for different data-sets, both the \optimal and the \bsqui perform well, regardless of the data-set used.
The \bsqui achieved similar \nma compared to the \optimal, while only needing polynomial runtime to be generated.
We conclude, that random forests are well suited as an anytime algorithm and can be used with the \bsqui, which can confidently and efficiently find a step order with a high mean accuracy.

\bibliographystyle{plain}
\bibliography{literature}

\end{document}